\documentclass{article}

\usepackage[english]{babel}

\usepackage[letterpaper,top=2cm,bottom=2cm,left=3cm,right=3cm,marginparwidth=1.75cm]{geometry}

\usepackage{amsmath}
\usepackage{graphicx}
\usepackage{booktabs}
\usepackage[numbers]{natbib}
\usepackage{authblk}
\usepackage[toc,page]{appendix}
\usepackage{longtable}
\usepackage[colorlinks=true, allcolors=blue]{hyperref}

\title{Early Diagnosis of Chronic Obstructive Pulmonary Disease from Chest X-Rays using Transfer Learning and Fusion Strategies}

\author[1]{Ryan Wang}
\author[1]{Li-Ching Chen}
\author[2]{Lama Moukheiber}
\author[2]{Mira Moukheiber}
\author[2]{Dana Moukheiber}
\author[3]{Zach Zaiman}
\author[ ]{Sulaiman Moukheiber}
\author[4]{Tess Litchman}
\author[4]{Kenneth Seastedt}
\author[5]{Hari Trivedi}
\author[6]{Rebecca Steinberg}
\author[1]{Po-Chih Kuo}
\author[5]{Judy Gichoya}
\author[2,4]{Leo Anthony Celi}

\affil[1]{Department of Computer Science, National Tsing Hua University}
\affil[2]{Institute for Medical Engineering and Science, Massachusetts Institute of Technology}
\affil[3]{Department of Computer Science, Emory University}
\affil[4]{Department of Medicine, Beth Israel Deaconess Medical Center}
\affil[5]{Department of Radiology, Emory University}
\affil[6]{Department of Medicine, Division of General Medicine, Emory University School of Medicine}

\date{30 September 2022}

\begin{document}
\maketitle
\pdfoutput=1

\begin{abstract}
Chronic obstructive pulmonary disease (COPD) is one of the most common chronic illnesses in the world and the third leading cause of mortality worldwide but is often underdiagnosed. COPD is difficult to diagnose early and often is not diagnosed until later in the disease course.  Spirometry tests are the gold standard for diagnosing COPD but can be difficult to obtain, especially in resource-poor settings such as low-income countries. Chest X-rays (CXRs), however, are readily available and may serve as a screening tool to identify patients with COPD who should undergo further testing. Currently, no research applies deep learning (DL) algorithms that use large multi-site data and multi-modal data to detect early-stage COPD patients, and this is the first study to address this problem using data fusion and model fusion techniques while also evaluating fairness across demographic subgroups. We, therefore, sought to develop DL algorithms to detect COPD using CXRs to create a screening tool for patients where spirometry may be difficult to access. We use three CXR datasets in our study, CheXpert to pre-train models, MIMIC-CXR, and Emory-CXR to develop and validate our models. The CXRs from patients in the early stage of COPD and not on mechanical ventilation are selected for model training and validation. We apply transfer learning for training on the base models and found the Xception base model performs better than other base models, including DenseNet121, ResNet50V2, and MobileNetV2. We visualize the Grad-CAM heatmaps of the true positive cases on the Xception base model for both MIMIC-CXR and Emory-CXR test datasets. We further propose two fusion schemes, (1) model-level fusion, including bagging and stacking methods using MIMIC-CXR, and (2) data-level fusion, including multi-site data fusion using MIMIC-CXR and Emory-CXR, and multi-modal fusion using MIMIC-CXRs and MIMIC-IV EHR, to improve the overall model performance. Fairness analysis is performed to evaluate if the fusion schemes have a discrepancy in the performance among different demographic groups (race-ethnicity, sex, and age). The model-level and multi-modal fusion schemes perform consistently on the MIMIC-CXR internal test data, and the multi-site data fusion scheme improves the model generalizability on the Emory-CXR test data compared to the other fusion strategies. The results demonstrate that DL models can detect COPD using CXRs only which can facilitate early screening, especially in low-resource regions where CXRs are more accessible than spirometry. Further studies on using CXR or other modalities to predict COPD ought in future work. All codes for the study are available at this link: \url{https://github.com/Ryan-RE-Wang/COPD-project} and the supplementary can be found at \url{https://docs.google.com/document/d/1N0VJtglD5DvtwbqjYiH8bdvt_LxNnswt3ZtzDyZpj1A/edit?usp=sharing}.
\end{abstract}

\section{Introduction}

Chronic obstructive pulmonary disease (COPD) remains one of the leading causes of mortality and morbidity \cite{mannino_global_2007}. Early intervention can mitigate disease progression and decrease healthcare expenditure \cite{decramer_treatment_2010, moretz_development_2015}. However, due to the lack of routine screening, COPD is still widely underdiagnosed \cite{hill_prevalence_2010}. In clinical practice, pulmonary function tests (PFT) and spirometry tests are necessary diagnostic tools for COPD diagnosis \cite{vogelmeier_global_2017}. Nevertheless, spirometry tests may fail to capture the presence of early COPD before it becomes detectable. Therefore asymptomatic patients are less likely to be tested \cite{andreeva_spirometry_2017,labonte_undiagnosed_2016,us_preventive_services_task_force_uspstf_screening_2016}. Further, the current screening tools, such as spirometry, are expensive and scarce in low-to-middle-income countries, leading to a delay in diagnosis \cite{beran_burden_2015,meghji_improving_2021}. Instead, the chest radiograph (CXR) is ubiquitous and cheap \cite{mettler_patient_2020}. The ability to create an early diagnostic screening tool for COPD from CXRs offers a significant clinical opportunity in developing a tool for COPD detection and thus targeting individuals for early interventions such as lung cancer screening and smoking cessation programs.

Recent studies have demonstrated the ability of deep learning (DL) models, such as convolutional neural networks (CNN), to classify radiographic findings from CXRs and achieve human-like performance \cite{lakhani_deep_2017,rajpurkar_chexnet_2017,tang_automated_2020,wang_lung_2017}, while also processing large amounts of images at high speed. Previous studies have mostly focused on classifying and detecting the 14 common chest radiographic findings seen on CXRs obtained from radiology reports using publicly sourced databases \cite{rajpurkar_chexnet_2017,rubin_large_2018,Wang_2017,yao_learning_2018}. Despite the success of DL in pulmonary disease classification using CXRs, a very limited number of studies have explored the potential of DL techniques in COPD diagnosis using CXRs only.

To the best of our knowledge, there is little attempt to develop a COPD prognostic model to identify the presence of COPD in early stages using both CXR and EHRs. In our study, we propose to build deep learning techniques using model-level and data-level fusion strategies to diagnose COPD using real-world data such as CheXpert, Medical Information Mart for Intensive Care (MIMIC-IV)\cite{johnson_alistair_mimic-iv_nodate}, MIMIC-CXR-JPG \cite{johnson_mimic-cxr-jpg_2019} and Emory data. Identifying COPD from CXRs and EHRs for early COPD screening could prompt early diagnosis, prevention, and treatment.\\

Our contributions are three-fold:
\begin{itemize}
    \item We implement model-level fusion strategies by combining four base models, including DenseNet121, ResNet50V2, Xception, and MobileNetV2 to build a primary screening model for COPD using real-world CXRs. To our knowledge, we are among the first to focus on model building for early screening of COPD.
    \item We use two data-level fusion strategies, including a multi-modal data strategy. (MIMIC-CXRs and MIMIC-IV EHR) to incorporate relevant clinical factors with CXRs and multi-site data strategy (MIMIC-CXRs and Emory CXRs) to improve the performance and generalizability of the base models.
    \item We assess model bias by evaluating the model performance of the data-level and model-level strategies across different demographic subgroups, including race-ethnicity, sex, and age.
\end{itemize}

\section{Related Work}
\subsection{COPD diagnostic models}
DL strategies, such as convolutional neural network (CNN) classifiers, have become a dominant approach in classifying radiographic findings from CXRs. Recent studies have sought to adapt various DL techniques to predict COPD; however, these studies present various limitations. They are limited to a very small number of COPD patients in both the training and testing set, only one data type \cite{pyrros_validation_2022}, and a single-institution \cite{schroeder_prediction_2021}. Further, the studies are limited in their consideration of the mechanical ventilation status of patients and thus pose uncertainty about the stage of COPD for early screening.

 \cite{degrave_ai_2021} demonstrated that the artificial intelligence system would learn spurious features as shortcuts to aid their classification rather than learning pathologically relevant features. For this concern, in \cite{schroeder_prediction_2021}’s study, the intubation may become a shortcut for models to determine the stage of COPD because the intubation would be obvious in the CXRs and indicative of respiratory status. To the best of our knowledge, the latent ability to recognize patients with early COPD and without mechanical ventilation solely depending on CXRs has not yet been investigated.

\subsection{Model-level and data-level fusion}
Two types of fusion strategies have shown to be promising in prediction tasks. Model-level fusion \cite{du_combining_2019} is one approach which combines base-model predictions to form a composite prediction \cite{ahmad_deep_2021,cao_ensemble_2020, ganaie_ensemble_2022, ragab_deep_2022} and thus leverages the strength of each model. Bagging and stacking are two popular model-fusion techniques \cite{ganaie_ensemble_2022}. Bagging integrates a set of predictions from several independent base models into a single prediction using an averaging or majority voting scheme. Stacking is an approach where the meta-model learns how to combine the output of the base models best to provide an optimal prediction. Model-level fusion schemes have been implemented in different medical imaging tasks and have been shown to perform better and more stable than single model architectures \cite{ahmad_deep_2021,barragan-montero_artificial_2021,chandra_coronavirus_2021,jia_atlas_2018,regan_clinical_2015,sirazitdinov_deep_2019}.

Data-level fusion schemes integrate multi-modal features or multi-site data into a single predictor \cite{huang_fusion_2020, huang_multimodal_2020, tariq_patient-specific_2021}. Various studies have demonstrated the consistent improvement in model performance using data-level fusion strategies \cite{huang_fusion_2020,huang_multimodal_2020,nam_deep_2022,tariq_patient-specific_2021, schroeder_prediction_2021}.
With the proliferation of publicly available medical imaging datasets and electronic health records (EHR), it is important to utilize pixel-level data and clinical patient data to obtain better feature representations during training. It is also important to use data from multiple centers to account for variability in data acquisition and processing to increase the validity of the proposed fusion method \cite{flynn2009benefits}. 

\subsection{Fairness evaluation across different subgroups}
Fairness is a rising concern in DL applications in the medical field. \cite{seyyed2021underdiagnosis} demonstrated that the state-of-the-art DL model consistently underdiagnose under-served patients. In \cite{seyyed2020chexclusion}'s study, the researchers found disparities in true positive rate (TPR) for pulmonary disease diagnosis across different demographic subgroups. Because TPR requires a binary threshold which would significantly affect the calculation of TPR, we use AUC as our fairness evaluation metric instead. 

Yet, there are only a few studies \cite{https://doi.org/10.48550/arxiv.2205.13421} investigating the evaluation of DL model on different patient subgroups and comparing it using real-world multimodal data. Our work aims to asses bias across different demographic groups to examine if our models yield biased prediction results. We evaluate the models across different demographic subgroups and calculate the AUC's standard deviation (SD) for each, respectively, to quantify the discrepancies in the AUC across the models. We further compare the SD of the AUC for the Xcpetion base model to that of the fusion strategies to see if the fusion strategies would decrease the SD of the AUC in order to mitigate the discrepancies across the models.

\section{Method}
\label{math}
\subsection{Data} 
\subsubsection{Chest X-Ray Datasets}
This study uses frontal CXRs from three datasets: CheXpert, MIMIC-CXR-JPG, and Emory-CXR. CheXpert is a retrospective dataset from the Stanford Hospital consisting of 224,316 CXRs of 65,240 patients \cite{irvin_chexpert_2019}. MIMIC-CXR-JPG is a publically available dataset containing 377,110 JPG images and its structured labels corresponding to 227,835 radiographic studies sourced from the Beth Israel Deaconess Medical Center between 2011 and 2016 \cite{goldberger_physiobank_2000,johnson_mimic-cxr_2019, johnson_mimic-cxr-jpg_2019}. Both CheXpert and MIMIC-CXR are labeled with the same 14 cardiopulmonary disease labels. Emory-CXR is a private dataset collected from five hospitals across the Emory Healthcare system between 2019 and 2020. This dataset is acquired from inpatient and outpatient hospitals and contains 226,640 CXRs from 57,909 patients. 
\subsubsection{Electronic Health Record (EHR) databases}
We use the MIMIC-IV database (version: 1.0), which is a publicly available EHR database maintained by Beth Israel Deaconess Medical Center from 2008 to 2019, consisting of more than 200,000 emergency department (ED) admissions and more than 60,000 intensive care units (ICU) stays \cite{goldberger_physiobank_2000,johnson_alistair_mimic-iv_nodate}. We also use the private EHR from Emory's institutional database, collected from five hospitals across the Emory Healthcare system.

\begin{table*}[htbp]
\centering
\small
{\vspace{-0.5cm}}
  {\caption{\label{tab:basic}Summary of the datasets used for the COPD prediction experiments.}}%
  {\begin{tabular}{llll}
  \toprule
  \bfseries & \bfseries MIMIC & \bfseries Emory & \bfseries CheXpert\\
  \midrule
  Number of patients (number of images) & 52,804 (201,748) & 57,909 (226,640) & 48,285 (140,921) \\
  \midrule
  Asian & 1,930 (3.7\%) & 2,035 (3.5\%) & 7,422 (15.4\%) \\
  Black & 8,923 (16.9\%) & 26,729 (46.2\%) & 3,016 (6.2\%) \\
  Latino & 3,340 (6.3\%) & 1,006 (1.7\%) & 2,322 (4.8\%) \\
Others & 4,720 (8.9\%) & 3,199 (5.5\%) & N/A \\
White & 33,891 (64.2\%) & 24,940 (43.1\%) & 34,657 (71.8\%) \\
\midrule
Female & 27,473 (52\%) & 30,236 (52.2\%) & 21,603 (44.7\%) \\
Male & 25,331 (48\%) & 27,673 (47.8\%) & 26,182 (55.3\%) \\
\midrule
0-40 & 8,216 (15.6\%) & 11,691 (20.2\%) & 5,880 (12.4\%)\\
40-60 & 16,433 (31.1\%) & 16,909 (29.2\%) & 13,565 (28.6\%)\\
60-80 & 19,493 (36.9\%) & 22,343 (38.6\%) & 18,339 (38.7\%)\\
80+ & 8,662 (16.4\%) & 6,966 (12\%) & 9,633 (20.3\%)\\
\bottomrule
  \end{tabular}}
\end{table*}

\subsubsection{Patient Cohort and Data Pre-processing}
We use the MIMIC-IV database to extract the patient cohort with COPD and those without COPD from both the ED and ICU based on the ICD diagnoses. We exclude images from patients who were on mechanical ventilation for two reasons. First, we did not want the algorithm to learn from the CXRs that patients intubated with a mechanical ventilation device are indicative of COPD. Second, our focus is to build an early-screening tool for patients  with COPD, and so we exclude COPD patients who are on mechanical ventilation as they are more likely to have an advanced disease. We used images from patients on room air, oxygen, or high-flow nasal cannula. We create binary labels for the CXRs using the ICD-9 and ICD-10 diagnosis codes from the MIMIC-IV and Emory EHR listed in Supplementary Table A1 to indicate the presence or absence of COPD. 

The patient's demographic extracted from MIMIC-IV and Emory EHR includes self-reported race-ethnicity, sex, and age. The summary of the datasets we use in our experiments is shown in Table~\ref{tab:basic}. All images are processed with histogram equalization, resized to (256, 256). The image pixel values are normalized from 0 to 1. We split the imaging data into training (64\%), validation (16\%), and testing (20\%) based on the patient's ID, and we ensure that patients in the training set are not included in the validation and the testing set to avoid data leakage.

\subsection{Transfer learning}
This study uses four CNN-derived state-of-the-art models, including DenseNet121, ResNet50V2, MobileNetV2, and Xception equipped with ImageNet pre-trained weights obtained from the TensorFlow Keras library. In the pre-training stage, we train the four base models to classify six radiographic labels using the CheXpert dataset to allow the models to learn pulmonary features. We select top six radiographic labels, including 'Atelectasisa', 'Cardiomegaly', 'Edema', 'Lung Opacity', 'Pleural Effusion', and 'Support Devices', out of fourteen common labels based on their prevalence in the CheXpert dataset. In the fine-tuning stage, each pre-trained model is fine-tuned on the MIMIC-CXR dataset for the COPD detection task. To accelerate the fine-tuning process, we freeze the learned weights for the first 30\% of the layers and fine-tune the remaining layers. To address the imbalance in the number of cases of COPD, we use the class weights parameter to give different weightage to the COPD and non-COPD classes. Since the class weights regularize the loss function, the model would give a higher weight for the loss of the COPD class while a lower weight for the loss of the non-COPD class. We calculate the class weights for the COPD and non- COPD class, which are defined as follows:

\begin{equation}
w_{COPD}=\frac{\text{total \# of CXRs}}{2*(\text{\# of COPD CXRs)}}
\end{equation}
\begin{equation}
w_{non-COPD}=\frac{\text{total \# of CXRs}}{2*(\text{\# of non-COPD CXRs)}}
\end{equation}

\begin{figure*}[htbp]
\centering
\caption{\label{fig:overview}Overview of our proposed framework.}
{\includegraphics[scale=0.35]{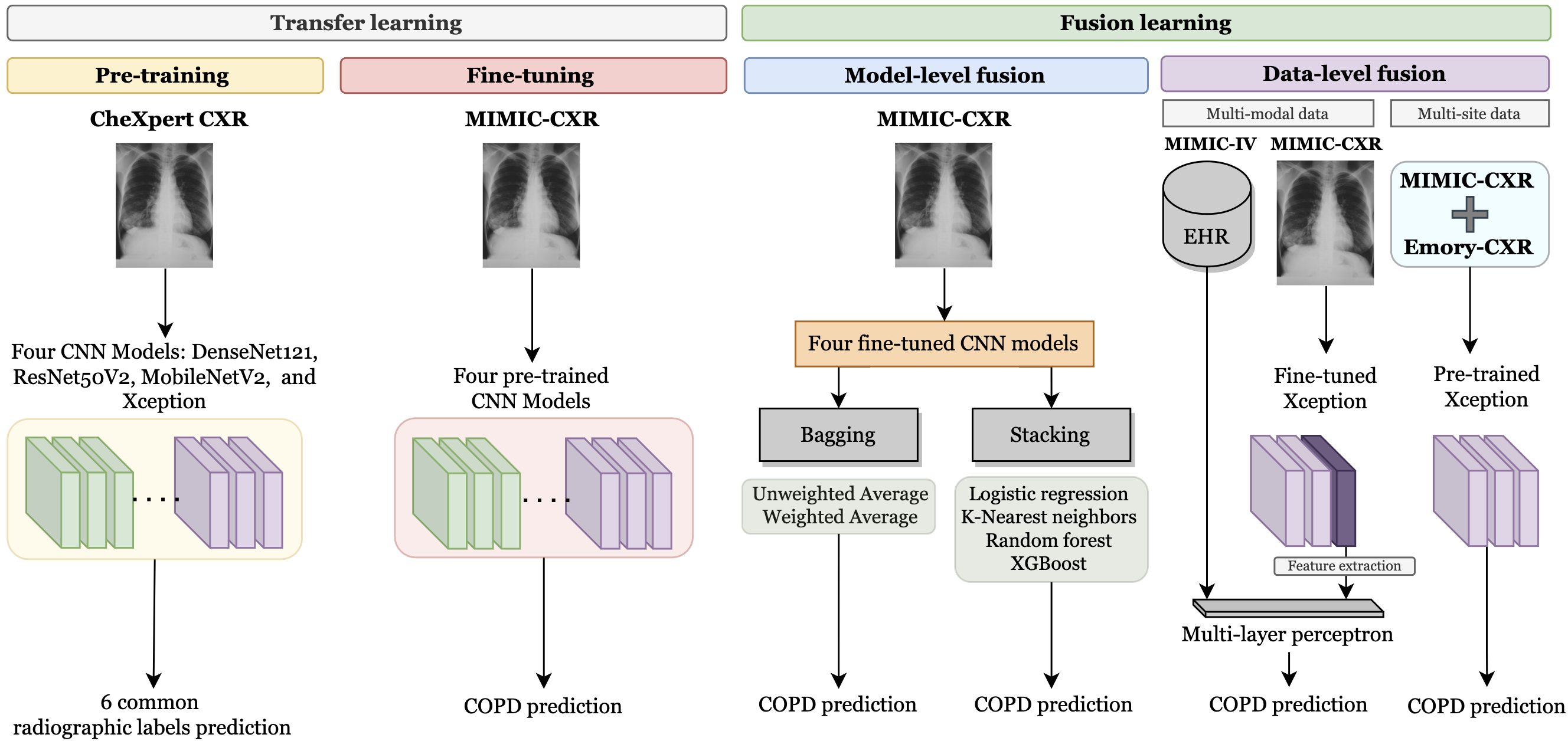}}
\end{figure*}

We apply 1.91 and 0.68 weights to the loss of the COPD and non-COPD classes, respectively. For the pre-training and fine-tuning stages, the learning rate is set to 0.001 and  decays 5\% for every epoch, and the batch size is set to 32. Adam optimizer is used, and binary cross entropy is used as the loss function. We stop the model training when the validation loss does not improve for three consecutive epochs. All hyperparameters and random seeds are identical for all training processes to ensure reproducibility.
\subsection{Fusion strategies}
To improve the performance of the fine-tuned models, we use the fine-tuned models as the base models to implement model-level fusion and data-level fusion strategies. In terms of model-level fusion, we implement bagging and stacking methods. For the bagging methods, we implement unweighted and weighted averaging. The stacking method contains the meta-models trained using the predictions of the base models. For data-level fusion, we experiment with two different techniques. First, we merge the MIMIC-CXRs and Emory-CXRs to create a multi-site data fusion strategy, which we use as our training set to train the best-performing base model (Xception). Second, we implement a multi-modal data fusion strategy ( joint fusion type II) \cite{huang_fusion_2020}, where we concatenate the feature representations of the CXRs from the penultimate Xception model layer with demographic variables from EHR. Our proposed approaches are shown in Figure~\ref{fig:overview}.

\subsubsection{Model-level fusion}
\paragraph{Unweighted average bagging}
We average the predictions of the four fine-tuned base models with equal weighting to obtain the final predictions. We later evaluate whether the unweighted average bagging strategy outperform the single base models.

\begin{figure}
\centering
\caption{\label{fig:dend}Dendrogram showing pairwise cosine distance for predictions across the four base models.}
 {\includegraphics[scale=0.6]{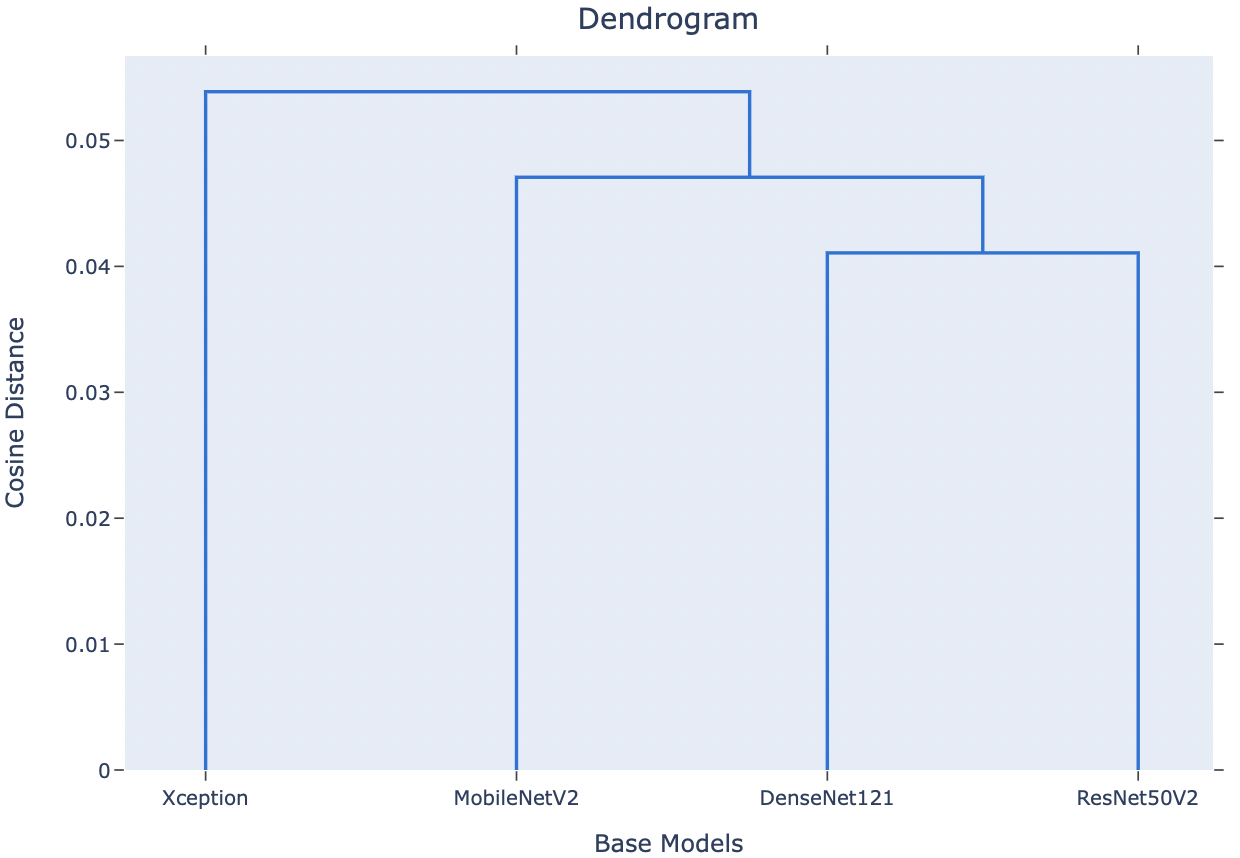}}
\end{figure}

\paragraph{Weighted average bagging}
We calculate the cosine distance of the validation set predictions obtained from the four base models and construct a dendrogram using the Unweighted Pair Group Method with an Arithmetic mean (UPGMA) algorithm, an hierarchical clustering approach \cite{2020SciPy-NMeth} defined as follows:
\begin{equation}
d(u, v)= \sum_{ij}{}\frac{d(u[i], v[i])}{|u|*|v|}
\text{,}
\end{equation}
where $|u|$ and $|v|$ are the cardinalities of clusters u and v, respectively for all points i and j. 
We average the test predictions of the base models based on the dendrogram as shown in Figure~\ref{fig:dend} to obtain the final prediction. For example, in Figure~\ref{fig:dend}, the predictions of the DenseNet121 and the ResNet50V2 are averaged first, followed by the predictions of the MobileNetV2, and finally, the predictions of the Xception. This is equivalent to assigning weights of  the predictions of the DenseNet121, ResNet50V2, MobileNetV2, and Xception as \( \frac{1}{8} \), \( \frac{1}{8} \), \( \frac{1}{4} \), and \(\frac{1}{2}\), respectively. We give the prediction of the base model a larger weight since the length of the dendrogram leg for the base model is longer.

\paragraph{Base model stacking}
We use the training data predictions of the base models as inputs to train the meta-model. The meta-model would optimize predictions based on the test data predictions from the base models. We implement four machine learning models as the meta-models, including logistic regression, k-nearest neighbors, XGBoost, and random forest.

\subsubsection{Data-level fusion}
\paragraph{Multi-site data fusion}
We merge two datasets, MIMIC-CXR and Emory-CXR, and use the combined dataset to fine-tune and evaluate the best-performing base model (Xception). We examine if data from different sources enhances the performance and generalizability of the multi-site data fusion strategy.

\paragraph{Multi-modal data fusion}
We construct a data-level fusion strategy with the CXRs and their corresponding EHRs as input. We use an Xception base model and a multi-layer perceptron model to create a multi-modal strategy (joint fusion—type II) \cite{huang_fusion_2020}. The learned features representations of the CXRs from the penultimate fine-tuned Xception layer are encoded into 64 hidden nodes concatenated with ten input nodes representing the demographic variables —five for race-ethnicity (Asian, Black, Hispanic, Others, and White), four for age (0-40, 40-60, 60-80, and 80+), and one for sex (Female or Male). 
 \subsection{Evaluation metrics} 
We report the point estimates and the 95\% confidence interval (CI) of the evaluation metrics: area under the receiver operating characteristic curve (AUC), precision, recall, F1-score, the area under the precision-recall curve (AUPRC), and balanced accuracy \cite{brodersen_balanced_2010} for all the experimented and proposed methodologies. 95\% (CI) is calculated by bootstrapping the metric scores over 1000 runs.
\subsection{Model interpretation} 
To understand which CXR regions the base models focus on, we visualize the hot spots for the true positive by using the Gradient-weighted Class Activation Mapping (GradCAM) algorithm \cite{selvaraju_grad-cam_2017}.

\subsection{Fairness analysis}
To examine whether there is a discrepancy in the models performance across demographic subgroups, we evaluate the average AUC and the standard deviation (SD) of the AUCs among different race-ethnicity, age groups, and sex separately on the MIMIC and Emory datasets. That is, the model would have separate results for each demographic group. The average AUC shows how the models perform in the demographic group, and the SD shows the deviation from the average AUC across demographic subgroups. 

\section{Results}
\label{result}
\subsection{Transfer learning}
\label{transfer}
We present the results of four different state-of-the-art CNN models tested on MIMIC-CXR and Emory-CXR in Table~\ref{tab:Base models tested on MIMIC-IV data} and Table~\ref{tab:Base models tested on Emory data} of Appendix~\ref{appendix:Basemodel}. The Xception base model yields the best AUC score (0.76), F1-score (0.54), AUPRC (0.54), and balanced accuracy (0.69) during the internal testing on MIMIC-CXR; however, the performance of the four base models is relatively similar. Further, when performing external testing on Emory-CXR, the Xception base model attains the best performance. Additionally, the performance of the Xception base model on both MIMIC-CXR and Emory-CXR is very similar, demonstrating that our fine-tuned model does not overfit on the training data. The detailed results with precision and recall are listed in Supplementary Table B1 to B2.

\subsection{Fusion strategies}
\label{fusion}
We present the model testing results of our fusion strategies in Table~\ref{tab:Fusion strategies tested on MIMIC data} for MIMIC-CXR and Table~\ref{tab:Fusion strategies tested on Emory data} of Appendix~ \ref{appendix:second} for Emory-CXR. For testing on MIMIC-CXR, the unweighted and weighted bagging models have the best AUC score (0.77), F1-score (0.54), AUPRC (0.56), and balanced accuracy (0.70). Since the Xception model performs the best among the base models, we use the Xception model in the data-level fusion strategy. The stacking model using the random forest as a meta-model performs the best among the other stacking models; however, it does not outperform the Xception basemodel. 

The multi-site data-level fusion strategy results show that merging the Emory-CXR dataset with MIMIC-CXR for training does not improve the performance when testing on MIMIC-CXR. Nonetheless, it performs better when testing on Emory-CXR than the Xception base model. The multi-modal data-level fusion strategy results show that incorporating demographic features does not improve the performance of both testing datasets. We use the chi-square test to evaluate the independence of the race-ethnicity, age, sex variables, and the 14 radiographic labels between the COPD and non-COPD cohorts. A lower p-value ($p<0.01$) demonstrates that the distributions of these variables are statistically significant between the COPD and non-COPD patient cohorts. The detailed results with precision and recall are listed in Supplementary Table B3 to B4.

\begin{table*}[htbp]
\centering
\small
{\caption{\label{tab:Fusion strategies tested on MIMIC data}Evaluation metrics averaged over 1000 epochs ±95\% for the various fusion strategies on MIMIC-CXR test data compared to the Xception base model. To validate our implementation, we shuffled the images' labels and used the randomly labeled images to train a model as a permutation test. The results of the permutation test indicated our implementation is sound.}}
{\begin{tabular}{llllll}
\toprule
\bfseries Fusion Strategies&\bfseries AUC&\bfseries F1-score&\bfseries AUPRC&\bfseries Balanced Accuracy \\
\midrule
Base Xception Model &
0.76 [0.75-0.76]&
0.53 [0.52-0.54]&
0.54 [0.53-0.55]&
0.69 [0.68-0.70]\\
\midrule
\bfseries Base Models Bagging &&&&\\
Unweighted Average &
0.77 [0.76-0.77]&
0.54 [0.53-0.55]&
0.56 [0.55-0.57]&
0.70 [0.69-0.70]\\
Weighted Average &
0.77 [0.76-0.77]&
0.54 [0.53-0.55]&
0.56 [0.55-0.57]&
0.70 [0.69-0.70]\\
\midrule
\bfseries Base Models Stacking &&&&\\
Logistic Regression&
0.76 [0.75-0.76]&
0.53 [0.52-0.54]&
0.54 [0.53-0.55]&
0.69 [0.68-0.69]\\
XGBoost&
0.75 [0.75-0.76]&
0.53 [0.52-0.54]&
0.54 [0.53-0.55]&
0.69 [0.68-0.69]\\
K-nearest Neighbors&
0.75 [0.74-0.75]&
0.52 [0.52-0.53]&
0.52 [0.51-0.53]&
0.68 [0.68-0.69]\\
Random Forest&
0.76 [0.76-0.77]&
0.54 [0.53-0.55]&
0.55 [0.54-0.56]&
0.69 [0.69-0.70]\\
\midrule
\bfseries Data-level Fusion &&&&\\
Multi-site&
0.75 [0.75-0.76]&
0.53 [0.52-0.54]&
0.54 [0.53-0.55]&
0.69 [0.68-0.69]\\
Multi-modal&
0.76 [0.75-0.77]&
0.54 [0.53-0.54]&
0.55 [0.53-0.56]&
0.69 [0.69-0.70]\\
\midrule
\bfseries Permutation test &
0.51 [0.50-0.52] &
0.31 [0.30-0.31] &
0.19 [0.18-0.19] &
0.51 [0.51-0.52]\\

  \bottomrule
  \end{tabular}}
\end{table*}

\newpage

\section{Discussion and Conclusion}
Our findings support our hypothesis that DL models trained using CXRs can identify patients with early-stage COPD. CXRs are a relatively accessible and affordable imaging modality. Therefore, our approach could provide a low-cost and widely available screening opportunity for COPD. We demonstrate that the Xception base model detects COPD with reasonable performance compared to the other base models as shown in Table~\ref{tab:Base models tested on MIMIC-IV data} and Table~\ref{tab:Base models tested on Emory data} of Appendix~ \ref{appendix:Basemodel}. The results are consistent for both MIMIC and Emory test datasets across the fusion strategies. The multi-site data-level fusion scheme perfroms better than the other schemes on the Emory-CXR test data indicating that our proposed strategy is generalizable. We use Grad-CAM to evaluate the Xception model's explainability. Looking at the true positive cases for both MIMIC-CXR and Emory-CXR datasets, we can see that the Grad-CAM of the Xception base model focuses on the upper right lung area.

The limitations of this study include reliance on ICD9/10 coding as labels and the lack of Pulmonary Function Testing (PFT) labels. PFT is an effective tool for the diagnosis of COPD. However, our study uses ICD codes to identify COPD diagnosis because concurrent PFT data was unavailable for many of the CXRs in our datasets. Instead, our study is utilized as a proof of concept that deep learning can be applied to patient data even if gold-standard PFT findings are not available. This has widespread implications because the current diagnostic criteria for COPD leads to a large population of underdiagnosed patients \cite{regan_clinical_2015,woodruff_clinical_2016}. 
Additionally, clinicians often only order PFTs when COPD progresses and becomes symptomatic, indicating that many early-stage diagnoses are often missed \cite{sood2016spirometric}. 
In the future, we will incorporate PFT findings to improve the performance of our model. We encourage hospitals to build their COPD screening model using their database based on our open-source code. Our proposed approach is advantageous for the COPD screening as it only requires CXRs, which are routinely collected and easily accessible.

\subsection{Explainability with Grad-CAM}
\label{gradcam}
We visualize the Grad-CAM heatmaps of the true positive cases on the Xception base model for both MIMIC-CXR and Emory-CXR test datasets. Table~\ref{tab:The heatmaps of the true positive cases} of \ref{appendix:ex_apd} shows the heatmaps of the true positive cases of the Xception model. We further average the heatmaps with predicted probabilities higher than 0.9 to obtain an averaged heatmap (Mean). We can see that the Xception base model mainly focuses on the upper right side of the lungs between the upper lung and the trachea, illustrating that COPD is associated with the upper airway \cite{Hurst2010-sx}.

\subsection{Fairness analysis}
\label{fair}
We evaluate the performance of the fusion strategies on each demographic group separately in both datasets. Table~\ref{tab:the fairness analysis of the fusion strategies on MIMIC-CXR dataset} and Table~\ref{tab:the fairness analysis of the fusion strategies on Emory-CXR dataset} show the results of the fairness analysis of each model on MIMIC-CXR and Emory-CXR. The average AUC scores show the average performance of the models among each demographic group, and the standard deviations show the performance gap between each demographic group. The star signs in the table represent the highest standard deviation, indicating the model's unfairness. The baseline and weighted average bagging models are the only two that do not have a star sign. That is, both models are reasonably fair across demographic groups. More importantly, since the weighted average bagging model has a higher average AUC, we can conclude that it is the best model that combines both reasonable performance and fairness evaluation across various demographic groups compared to other fusion strategies. The detailed results with precision and recall are listed in Supplementary Table B5 to B18.

\bibliographystyle{abbrv}
\bibliography{main}

\begin{appendices}

\section{Results of Base Models}
\label{appendix:Basemodel}
In this section, we list the performance metrics on the base models for both MIMIC-CXR and Emory-CXR data in \ref{transfer}.
\begin{table*}[htbp]
\centering
\small
  {\caption{\label{tab:Base models tested on MIMIC-IV data}Base models tested on MIMIC-CXR data.}}%
  {\begin{tabular}{lllll}
  \toprule
 \bfseries Model &\bfseries AUC&\bfseries F1-score&\bfseries AUPRC&\bfseries Balanced Accuracy \\
  \midrule
DenseNet121&
0.74 [0.74-0.75]&
0.52 [0.51-0.53]&
0.52 [0.51-0.53]&
0.68 [0.67-0.69]\\
MobileNetV2&
0.75 [0.74-0.76]&
0.53 [0.52-0.53]&
0.53 [0.52-0.54]&
0.68 [0.68-0.69]\\
ResNet50V2&
0.74 [0.73-0.75]&
0.52 [0.51-0.52]&
0.52 [0.51-0.53]&
0.67 [0.67-0.68]\\
Xception&
0.76 [0.75-0.76]&
0.53 [0.52-0.54]&
0.54 [0.53-0.55]&
0.69 [0.68-0.70]\\
  \bottomrule
  \end{tabular}}
\end{table*}
\begin{table*}[htbp]
\centering
\small
  {\caption{\label{tab:Base models tested on Emory data}Base models tested on Emory-CXR data.}}%
  {\begin{tabular}{lllll}
  \toprule
 \bfseries Model &\bfseries AUC&\bfseries F1-score&\bfseries AUPRC&\bfseries Balanced Accuracy \\
\midrule
DenseNet121
&0.73 [0.72-0.73]
&0.43 [0.42-0.44]
&0.40 [0.38-0.41]
&0.66 [0.66-0.67]\\
MobileNetV2
&0.72 [0.72-0.73]
&0.42 [0.41-0.43]
&0.39 [0.38-0.40]
&0.66 [0.65-0.67]\\
ResNet50V2
&0.72 [0.71-0.73]
&0.42 [0.41-0.43]
&0.39 [0.38-0.40]
&0.66 [0.65-0.67]\\
Xception
&0.74 [0.73-0.75]
&0.43 [0.43-0.44]
&0.40 [0.39-0.42]
&0.67 [0.67-0.68]\\
  \bottomrule
  \end{tabular}}
\end{table*}

\section{Results of Fusion Strategies}
\label{appendix:second}
We include the results of a different test dataset, MIMIC-CXR, for the same experiments presented in Table~\ref{tab:Fusion strategies tested on Emory data} of \ref{fair}.
\begin{table*}[htbp]
\centering
\small
  {\caption{\label{tab:Fusion strategies tested on Emory data}Evaluation metrics  averaged over 1000 epochs ±95\% CI for the various fusion strategies on Emory-CXR test data compared to the Xception base model.}}%
  {\begin{tabular}{lllllll}
  \toprule
\bfseries Fusion Strategies&\bfseries AUC&\bfseries F1-score&\bfseries AUPRC&\bfseries Balanced Accuracy\\
\midrule
Base Xception Model
&0.74 [0.73-0.75]
&0.43 [0.43-0.44]
&0.40 [0.39-0.42]
&0.67 [0.67-0.68]\\
\midrule
\bfseries Base Models Bagging &&&&\\
Unweighted Average &
0.75 [0.74-0.75]&
0.44 [0.43-0.45]&
0.42 [0.41-0.43]&
0.68 [0.67-0.68]\\
Weighted Average &
0.75 [0.74-0.75]&
0.44 [0.43-0.45]&
0.42 [0.41-0.43]&
0.68 [0.67-0.68]\\
\midrule
\bfseries Base Models Stacking &&&&\\
Logistic Regression
&0.74 [0.73-0.74]
&0.44 [0.43-0.45]
&0.41 [0.39-0.42]
&0.67 [0.66-0.68]\\
XGBoost
&0.73 [0.73-0.74]
&0.44 [0.43-0.45]
&0.40 [0.39-0.42]
&0.67 [0.66-0.68]\\
K-nearest Neighbors
&0.72 [0.72-0.73]
&0.43 [0.42-0.44]
&0.39 [0.38-0.40]
&0.67 [0.66-0.67]\\
Random Forest
&0.74 [0.74-0.75]
&0.44 [0.43-0.45]
&0.41 [0.40-0.43]
&0.68 [0.67-0.68]\\
\midrule
\bfseries Data-level Fusion &&&&\\
Multi-site&
0.76 [0.75-0.76]&
0.45 [0.44-0.45]&
0.43 [0.42-0.44]&
0.68 [0.68-0.69]\\
Multi-modal&
0.74 [0.73-0.74]&
0.44 [0.43-0.45]&
0.41 [0.39-0.42]&
0.67 [0.67-0.68]\\
  \bottomrule
  \end{tabular}}
\end{table*}

\onecolumn

\section{Additional Experimental Results}
\label{appendix:third}
\subsection{Explainability with Grad-CAM}
The model is highlighting areas commonly affected by COPD, such as the apex of the lung fields or the areas of the mediastinum affiliated with the pulmonary vasculature. The lung apices can often become emphysematous in COPD, and the pulmonary vessels may become more prominent as pulmonary hypertension worsens in the later stages of COPD. This is likely why the models are identifying these regions to make predictions. We will add the interpretation of the heatmap to the appendix so that the reviewers can understand the heatmap from a clinical perspective.
\label{appendix:ex_apd}
\begin{table*}[htbp]
\centering
\small
  {\caption{\label{tab:The heatmaps of the true positive cases}The first and second rows display the Grad-CAM heatmaps of the true positive cases on the Xception base model using the MIMIC-CXR test data and the Emory-CXR test data, respectively.}}%
  {\begin{tabular}{lllll}
  \toprule
 \hspace{0.35cm}\bfseries Pred: 0.996 &\hspace{0.35cm}\bfseries Pred: 0.996 &\hspace{0.35cm}\bfseries Pred: 0.994&\hspace{0.35cm}\bfseries Pred: 0.992&\hspace{0.8cm}\bfseries Mean\\
\midrule
{\includegraphics[width=75pt]{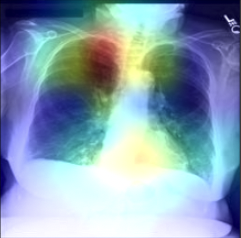}} & {\includegraphics[width=75pt]{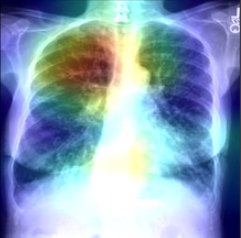}} &
{\includegraphics[width=75pt]{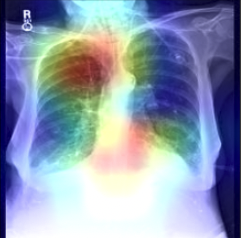}} & {\includegraphics[width=75pt]{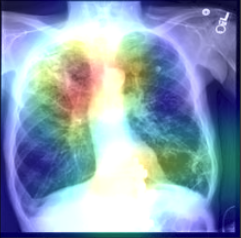}} & {\includegraphics[width=75pt]{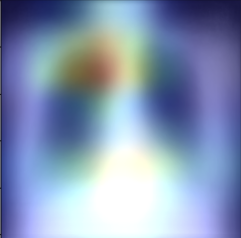}}\\
\midrule
\hspace{0.35cm}\bfseries Pred: 0.996 &\hspace{0.35cm}\bfseries Pred: 0.996 &\hspace{0.35cm}\bfseries Pred: 0.994&\hspace{0.35cm}\bfseries Pred: 0.992&\hspace{0.8cm}\bfseries Mean\\
 \midrule
{\includegraphics[width=75pt]{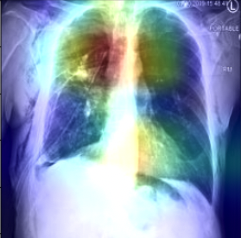}} & {\includegraphics[width=75pt]{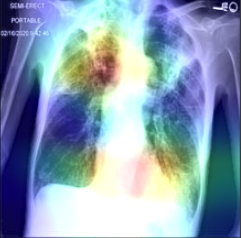}} &
{\includegraphics[width=75pt]{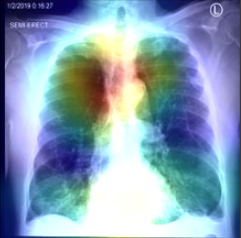}} & {\includegraphics[width=75pt]{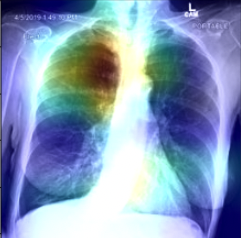}} & {\includegraphics[width=75pt]{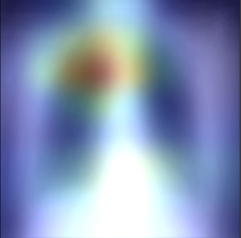}}\\
  \bottomrule
  \end{tabular}}
\end{table*}

\subsection{Fairness Analysis}
\label{appendix:fair_apd}
\begin{table*}[htbp]
\centering
\small
  {\caption{\label{tab:the fairness analysis of the fusion strategies on MIMIC-CXR dataset}Average AUC ±SD for model-level and data-level fusion strategies on MIMIC-CXR test dataset across various demographic subgroups.}}%
  {\begin{tabular}{llll}
  \toprule
\bfseries Fusion Strategies
&\bfseries Race-ethnicity
&\bfseries Sex
&\bfseries Age\\
\midrule
Base Xception Model&
0.78 [0.026]&
0.76 [0.010]&
0.73 [0.036]
\\\midrule
Unweighted Average Bagging&
0.79 [0.031]&
0.77 [0.010]&
0.74 [0.030]
\\ \midrule
Weighted Average Bagging&
0.79 [0.029]&
0.77 [0.010]&
0.74 [0.032]
\\ \midrule
Base Models Stacking (Random Forest Classifier)&
0.78 [0.030]&
0.77 [0.015] $\star$ &
0.73 [0.029]
\\ \midrule
Multi-site Data Fusion&
0.78 [0.025]&
0.76 [0.005]&
0.71 [0.046] $\star$
\\ \midrule
Multi-modal Data Fusion&
0.78 [0.032] $\star$ &
0.76 [0.010]&
0.72 [0.035]
\\ \bottomrule
  \end{tabular}}
\end{table*}

\begin{table*}[htbp]
\centering
\small
  {\caption{\label{tab:the fairness analysis of the fusion strategies on Emory-CXR dataset}Average AUC ±SD for model-level and data-level fusion strategies on Emory-CXR test dataset across various demographic subgroups.}}%
  {\begin{tabular}{llll}
  \toprule
\bfseries Fusion Strategies
&\bfseries Race-ethnicity
&\bfseries Sex
&\bfseries Age\\
\midrule
Base Xception Model&
0.72 [0.024]&
0.74 [0]&
0.70 [0.030]
\\\midrule
Unweighted Average Bagging&
0.72 [0.038]&
0.75 [0.005] $\star$ &
0.71 [0.025]
\\ \midrule
Weighted Average Bagging&
0.73 [0.030]&
0.75 [0]&
0.71 [0.027]
\\ \midrule
Base Models Stacking (Random Forest Classifier)&
0.71 [0.038]&
0.74 [0.005] $\star$ &
0.70 [0.029]
\\ \midrule
Multi-site Data Fusion&
0.74 [0.050] $\star$ &
0.76 [0]&
0.72 [0.031] $\star$
\\ \midrule
Multi-modal Data Fusion&
0.72 [0.038]&
0.74 [0]&
0.69 [0.026]
\\ \bottomrule
  \end{tabular}}
\end{table*}

\end{appendices}

\end{document}